\documentclass[10pt,conference]{IEEEtran}
\IEEEoverridecommandlockouts

\usepackage[utf8]{inputenc}
\usepackage[english]{babel}
\usepackage{fancyhdr}
\usepackage{lastpage}
\usepackage{placeins}

\usepackage{amsmath,amssymb,amsfonts}
\usepackage{algorithmic}
\usepackage{graphicx, subcaption}
\usepackage{textcomp}
\usepackage{xcolor}
\usepackage[backend=biber]{biblatex}
\def\BibTeX{{\rm B\kern-.05em{\sc i\kern-.025em b}\kern-.08em
    T\kern-.1667em\lower.7ex\hbox{E}\kern-.125emX}}

\begin{document}

\title{Top K Relevant Passage Retrieval for Biomedical Question Answering \\
}

\author{\IEEEauthorblockN{Shashank Gupta}
\IEEEauthorblockA{\textit{Department of Computer Science} \\
\textit{University of Kentucky}\\
shashank.gupta@uky.edu}

}
\maketitle


\section{Abstract}
Question answering (QA) is a task that answers factoid (what, when, where, etc.) questions using a large collection of documents. It aims to provide precise answers in response to the user's questions in natural language. Question answering relies on efficient passage retrieval to select candidate contexts, where traditional sparse vector space models, such as TF-IDF or BM25, are the de facto method. On the web, there is no single article that could provide all the possible answers available on the internet to the question of the problem asked by the user. The existing Dense Passage Retrieval model \cite{karpukhin2020dense} has been trained on Wikipedia dump from Dec. 20, 2018, as the source documents for answering questions.  Question answering (QA) has made big strides with several open-domain and machine comprehension systems built using large-scale annotated datasets. However, in the clinical domain, this problem remains relatively unexplored. According to multiple surveys, Biomedical Questions cannot be answered correctly from Wikipedia Articles. In this work, we work on the existing DPR framework for the biomedical domain and retrieve answers from the Pubmed articles which is a reliable source to answer medical questions. When evaluated on a BioASQ QA dataset, our fine-tuned dense retriever results in a \textbf{81\%} F1 score. 

\section{Introduction}
Question answering (QA), the process of identifying short and accurate answers to user questions written in natural language, is a long-standing topic that has been extensively studied in the open field over the past few decades. However,  in the field of biomedicine, most existing systems support a limited number of question-and-answer types and require further efforts to improve performance in terms of the accuracy of the supported questions. It's still a real challenge. 

Biomedical knowledge acquisition is an important task in information retrieval and knowledge management. Professionals, as well as the general public, need effective assistance to access, understand and consume complex biomedical concepts. For example, doctors always want to be aware of up-to-date clinical evidence for the diagnosis and treatment of diseases, and the general public is becoming increasingly interested in learning about their own health conditions on the Internet and using online search engines like Google search, Yahoo, etc. for advice or answers for their questions.

On the web, there is no single article that could provide all the possible answers available on the internet to the question of the problem asked by the user. Most search engines provide top results for an open-domain question but may also provide some biased articles which may or may not be directly relevant to the question asked by the user. These biased articles may be based on various criteria like advertised articles, paid articles to come on top results, the number of times a particular site is visited, and many more.

How does question-answering work on the Internet in general? The user writes a question into the search engine, and then search engines retrieve some top k passages where some of them contain the answer to the question from the huge corpus containing millions of documents D on the internet ($k \ll  $D).

Passage retrieval is becoming an important task in question answering because answering a question correctly really depends on these top k retrieved documents. If these top retrieved documents are not relevant to the question asked, the reader model (another deep learning model which finds the span of the token within the retrieved passages that may contain the answer) would not be able to find the span of the tokens which contains the answer and the user query would not be resolved.

By leveraging the now standard BERT pre-trained model \cite{devlin2019bert} and a dual-encoder architecture (Bromley et al., 1994), we focus on developing the right training scheme using a relatively small number of question and passage pairs. Through a series of careful ablation studies, our final solution is surprisingly simple: the embedding is optimized for maximizing the inner products of the question and relevant passage vectors, with the objective of comparing all pairs of questions and passages in a batch.

 In this work, we will be using a deep learning model, Dense Passage Retrieval (DPR) which uses BERT as encoders to extract questions and online article features to extract top relevant passages without any bias that could provide the answer from the PubMed abstracts to the medical question asked by the user.
 
\section{Motivation}

Acquiring biomedical information is pivotal for advanced knowledge management and sophisticated information retrieval. Both professionals and the broader populace require robust tools to navigate, comprehend, and assimilate intricate biomedical terminologies. For instance, medical practitioners consistently seek the latest clinical evidence to diagnose and treat ailments, while the general public is progressively leveraging online platforms such as Google Search, Yahoo, and others to gain insights into their health conditions and seek guidance.

Digitally, it's impractical to expect a singular article to encompass every conceivable answer to a user's inquiry. While most search engines prioritize top results for broad-domain queries, there's a potential risk of encountering articles with inherent biases. These biases might arise from factors such as promotional content, sponsored articles securing prime visibility, and frequent site visitations, among others.

The role of passage retrieval is gaining prominence in the realm of question-answering systems. The efficacy of a correct response is intrinsically tied to the relevance of the top 'k' documents retrieved. Should these documents lack pertinence to the posed question, the reader model—a deep learning construct designed to pinpoint answer spans within the retrieved content—may falter in identifying the correct token span, leaving the user's query unresolved. 

\section{Related Work}
In recent years, the scientific community has paid a lot of attention to question answering. The Text Retrieval Conference (TREC), which has been held annually since 1999, has benefitted open-domain quality assurance research. Otherwise, biomedical quality assurance has become difficult in recent years. Because of the intricacy of natural language in the biological sector, there has been no major advancement in this area. The primary integrated QA systems will be discussed in this section, with a special emphasis on the biomedical area.

In this regard, \cite{Lee2006BeyondIR} have created the MedQA medical quality assurance system which is composed of five components including (1) question classification, (2) query generation, (3) document retrieval, (4) answer extraction, and (5) text summarization. Despite the fact that the MedQA system gives brief summaries that might possibly answer medical inquiries, the system's present capability is limited: it can only answer definitional queries.

\textbf {\cite{article} developed HONQA}, a biomedical quality assurance system that takes phrases from HON-certified websites and offers them as responses to biomedical queries. In its current state, it is unable to deliver precise replies to question kinds such as yes/no and factoids.

\textbf{EAGLi, a biomedical QA system created by \cite{article1}}, intends to extract solutions to biological queries from MEDLINE records. Because it only covers definitional and factoid inquiries, the present EAGLi's capability is restricted to Wh-type queries.

\textbf{\cite{article3} proposed AskHERMES}, a clinical QA system that gives brief summaries as responses to ad hoc clinical queries. For all question kinds, the system only enables a single answer type in the form of multiple-sentence passages.

\textbf{\cite{Hristovski2015BiomedicalQA}} have introduced a biomedical QA system, \textbf{SemBT}, based on semantic relations extracted from the biomedical literature.

While the development of QA systems indicates progress since the start of the BioASQ challenge, the systems are still limited. Many were restricted to extremely particular circumstances and covered a narrow range of queries and responses. Furthermore, the majority of these solutions are not end-to-end QA systems. Our objective in this study is to go above past biological QA systems and create a QA system that can automatically handle a wide range of questions kinds such as yes/no questions, factoid questions, and so on.

\section{Problem Statement}

The problem of biomedical Question Answering (QA) in this project can be described as follows. Given a biomedical question, such as "How many genes constitute the DosR regulon, controlled by the dormancy survival regulator (DosR) in Mycobacterium tuberculosis?" or "What is the function of the gene MDA5?", our system is built to find a relevant top passages/contexts from a large corpus of diversified biomedical topics from which these questions can be answered. We assume in extracting such passages, each question can be answered by some spans in one or more passages in the corpus. As we can see, to answer a question correctly, it is important to extract the top passages efficiently. 

Passage retrieval is becoming an important task in question answering because answering a question correctly really depends on these top $k$ retrieved documents. If these top retrieved documents are not relevant to the question asked, the reader model (another deep learning model which finds the span of the token within the retrieved passages that may contain the answer) would not be able to find the span of the tokens which contains the answer and the user query would not be resolved. Hence the Retriever model plays the most important role in every QA system. 

Assume our dataset contains $X$ consisting of n tokens $d_1, d_2, d_3, d_4, d_5,..., d_D.$ We split each document into several passages, then we get a corpus with $M$ total passages such that corpus $C=$\{$p_1, p_2, p_3, p_4, p_5,..., p_M$\}. To answer a specific question, a QA system needs to include an efficient retriever component that can select a small set of relevant passages, before applying the reader to extract the answer, where the reader is used to extract answer spans from the selected passages. In our project, we are building an efficient retriever $R$, which can be applied to the corpus $C$, such that $R(q, C) \rightarrow{C_F}$, where $C_F$ is a subset of passages with $\lvert C_F\rvert=k$ $\ll$ $\lvert{C} \rvert$. For a fixed $k$, a retriever can be evaluated in isolation on top $k$ retrieval accuracy, which is the fraction of questions for which $C_F$ contains a span that answers the question. In this project, we will be using a deep learning model, Dense Passage Retrieval (DPR) which uses BERT as encoders to extract question and passage features.

\section{Overview}
Our dense passage retriever (DPR) employs a dense encoder EP(.), which encodes every text passage to a d- dimensional real-valued vector and creates an index for all M passages that will be retrieved.

At running time, DPR utilizes a different encoder EQ(.), which translates the input query to a d-dimensional vector and returns k passages comprising the vectors that are nearest to the query vector. Using the dot product of their vectors, we calculate the closeness here between the question and the passage:

\[ sim(q,p) = E_Q(q)^T E_P(p)\]

\begin{figure}[ht]
\label{Fig:Overview}

      \includegraphics[width=8cm, height=4cm]{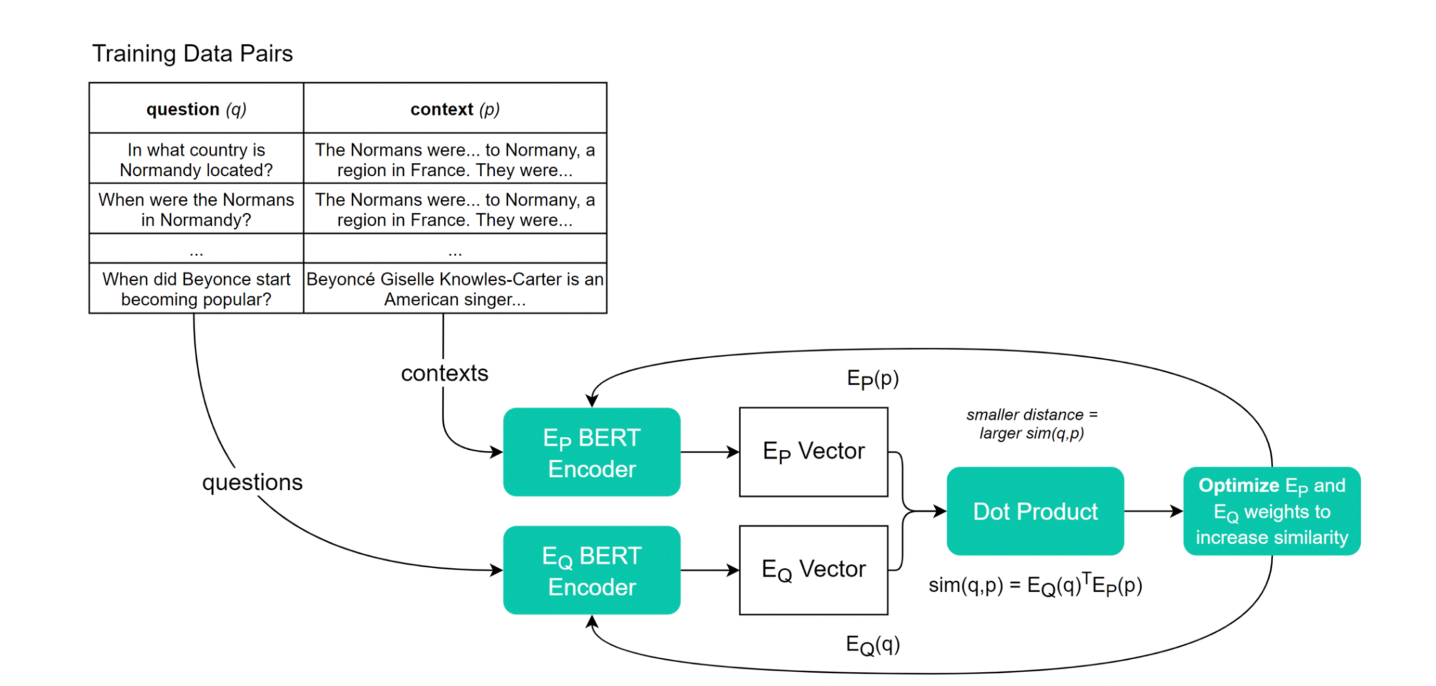}
      \caption{A schematic diagram illustrating training of DPR model}
\end{figure}

\section{Datasets}
We use the BioASQ 2021 biomedical QA dataset and training /dev/testing splitting method. Below we briefly describe the dataset and refer readers to their paper for the details of data preparation.

\textbf{BioASQ}: BioASQ 2021 task b is a question-answering dataset that contains around 3500 questions of types summary, list, factoids, and yes/no. Instances in the BioASQ dataset are composed of a question (Q), human-annotated exact answers (A) partial and exact both, and the relevant contexts (C).

\textbf{(1)Yes/no questions}: questions that require either a “yes” or “no” answer.

\textbf{(2)Factoid questions}: questions that require a particular entity name (e.g., of a disease, drug, or gene), a number, or a similar short expression as an answer. 

\textbf{(3)List questions}: questions that expect a list of entity names (e.g., a list of gene names, a list of drug names), numbers, or similar short expressions as an answer. 

\textbf{(4)Summary questions}: questions that expect short summaries as answer.

\section{Proposed Approach}
In this section, we describe the data we used for our approach and the basic setup.
\subsection{PubMed Data Pre-processing}
PubMed, a free search engine accessing primarily the MEDLINE database of references and abstracts on life sciences and biomedical topics, has more than 30 million citations and abstracts dating back to 1966. As of the same date, 20 million of PubMed's records are listed with their abstracts, and 21.5 million records have links to full-text versions. We use the PubMed dump from December 12, 2021, as the source documents for answering biomedical questions. Following [1], we extract the clean text portion of the PubMed articles. This step removes the tables, diagrams boxes, and lists. We then split each article into multiple, disjoint text blocks of 100 words as passages, serving as our basic retrieval units. Each passage is also prepended with the title of the PubMed article where the passage is from, along with a [SEP] token.

\subsection{Pre Processing BioASQ Dataset}
We downloaded BioASQ 2021 task b Question Answering dataset which contained around 3500 training and testing question answer pairs along with relevant context. Since we are finding the answers to the questions from the Pubmed abstracts, we replaced the relevant contexts of a question in the BioASQ dataset with the Pubmed abstracts by string matching and picking up the abstract which contains this context. We added some positive passages of questions as a negative passage to other questions and one hard negative.

\subsection{Models}
Since we are retrieving biomedical passages, we have used BioBERT, BERT base, and Facebook finetuned BERT as encoders for both questions and passages, compared to the general BERT used in [1]. We encode each question and passage to a high-dimension vector and perform maximum inner product search (MIPS), which can efficiently retrieve top $K$ passages for us. We use FAISS to index the dense representations of all passages. Specifically, we use IndexFlatIP for indexing and the exact maximum inner product search for queries.
\
\subsection{Positive and Negative Passages}
Since the original passage of the two datasets is pre-processed differently, we match and replace each positive passage with the passages in our split Pubmed passages. Passages found using the answer are also considered positive passages. We consider two different types of negatives: (1) Random: any random passage from the corpus; (2) BM25: top passages returned by BM25 which don’t contain the answer but match most question tokens (Hard Negatives).

\section{Training}

It is essentially a metric learning issue to train the encoders such that the dot-product similarity becomes a decent ranking function for retrieval. By learning a greater embedding function, the purpose is to develop a vector space wherein relevant pairings of questions and passages have a shorter distance (i.e., higher similarity) versus irrelevant ones.

\[D = ({q_i,p_i^+,p_{i,1}^-,...,p_{i,n}^- })_{i=1}^m\]
be the training data that consists of m instances. Each instance contains one question $q_i$ and one relevant (positive) passage $p_ i^+$, along with n irrelevant (negative) passages $p_{i,j}^-$. We optimize the loss function as the negative log-likelihood of the positive passage:

\[L({q_i,p_i^+,p_{i,1}^-,...,p_{i,n}^- })_{i=1}^m\]

\[-log \frac{e^{sim(q_i,p_i^+)}}{e^{sim(q_i,p_i^+)} + \sum_{j=1}^{n}{e^{sim(q_i,p_{i,j}^-)}}}\]
\subsection{PubMed Data Collection}
We extract the clean text portion of all 30 million PubMed articles 
from the NIH library and stored them on the LCC cluster provided by the University of Kentucky. 
\subsection{Question Answering Dataset}
We downloaded BioASQ 2021 dataset which contained 3742 questions and 500 golden sets for testing. We only selected question types of factoid and yes/no. We dropped summarization, and list type of questions. Hence our dataset had training (2057), dev (52), and testing dataset (102). The raw dataset is not in the format which is required for the input of your Dense Passage Retriever (DPR), hence prepossessing was required to align it with the DPR format as below. 

\subsection{Positive Passages}
Since the original positive context of the dataset for a question is pre-processed differently, we perform the string matching and replace each positive context with the articles from our collection of 30 million Pubmed abstracts. 

\subsection{Hard Negative}
During inference, the retriever needs to identify positive (or relevant) passages for each question from a large collection containing millions of candidates. However, during training, the model is learned to estimate the probabilities of positive passages in a small candidate set for each question, due to the limited memory of a single GPU (or other device). To reduce such a discrepancy, we tried to design specific mechanisms for selecting a few hard negatives from the top-k retrieved candidates. To find the hard negatives, we sub-sampled the 210K Pubmed articles randomly and ran the BM25 model to find the hard negatives on them. BM25 gives the score of how relevant a passage is given a query. The top passage which does not contain the answer was taken as a hard negative for the given query.

\subsection{In-Batch Negatives}
Assume that there are $B$ questions in a batch, and each question has one positive passage. With the in-batch negative trick, each question can be further paired with $B$-1 negatives (i.e., positive passages of the rest questions) without sampling additional negatives. We use the batch size of 16 which means each question has 1 positive context, 1 hard negative passage, and 15 negative passages.

\begin{figure}[ht]
      \includegraphics[width=8cm, height=3cm]{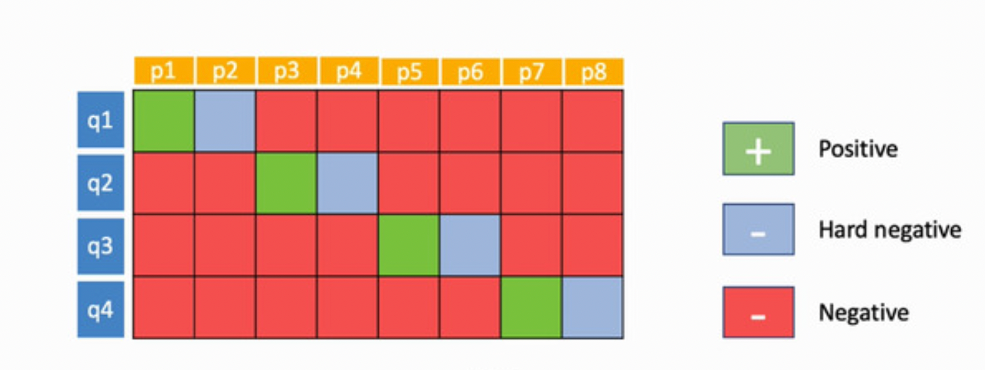}
      \caption{In-Batch Negatives}
\label{Fig:Negatives}

\end{figure}

\subsection{Architecture }
We use Haystack's version of DPR and used "Facebook dpr question encoder" for questions and "Facebook dpr ctx encoder" for contexts and fine-tuning these pre-trained Facebook encoders, compared to the general BERT used in [1].

\subsection{Training Setting}
We used 2057 questions from BioASQ to train our DPR model with a batch size of 16 and a total 8 number of epochs. 

\begin{figure}[ht]

      \includegraphics[width=9cm, height=4.5cm]{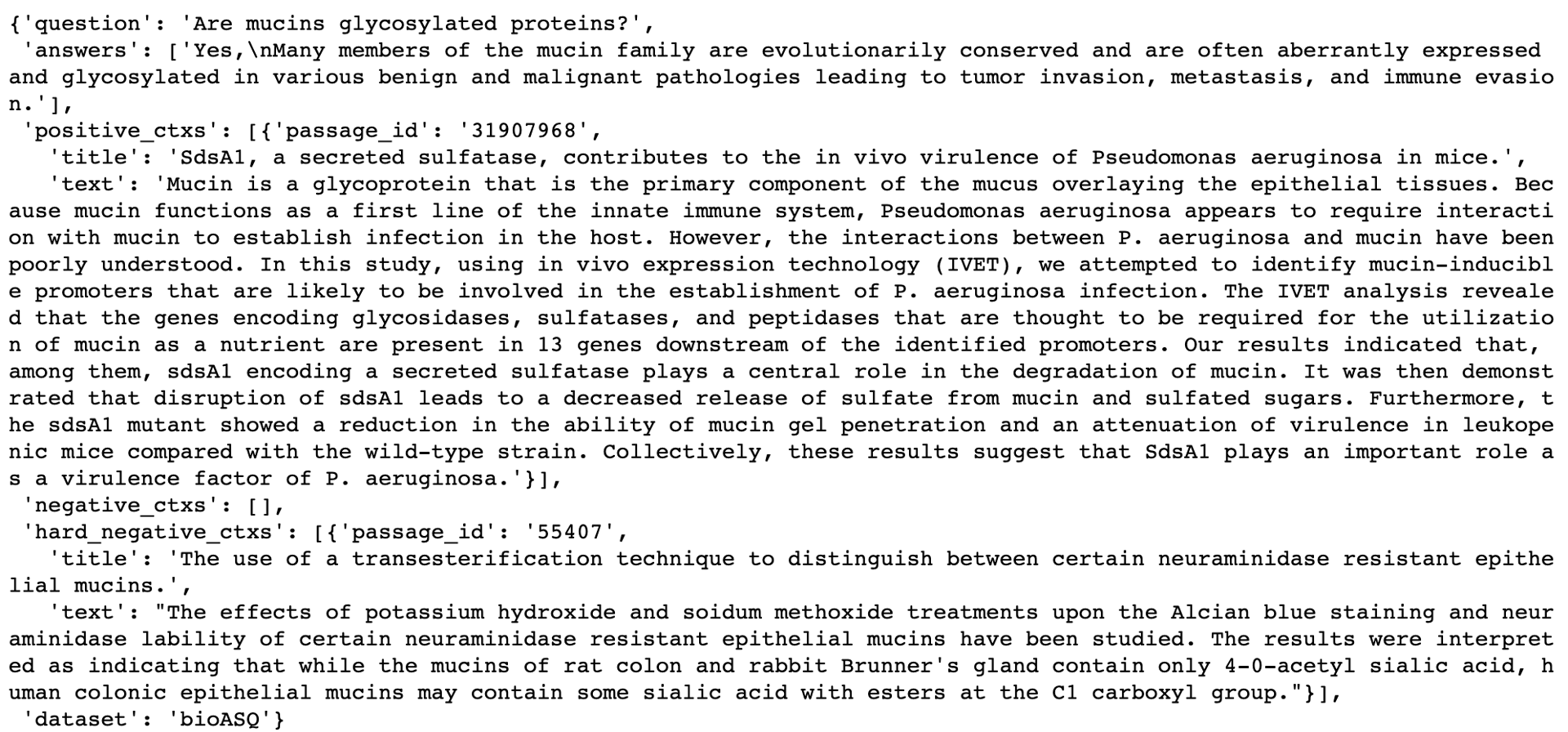}
      \caption{Dataset structure for the input to the model}
      \label{bioasq}
\end{figure}

The input to the model is shown in Fig. \ref{bioasq}.

We tried to experiment with the training of our model by using different passage and query encoders i.e. BioBERT v1.1, BERTbase, and Facebook pre-trained BERT. We also experimented with the number of epochs being 2,4 and 8. We fixed the batch size to 16.
\section{Results and observations}
\begin{table}[!h]
\label{T:equipos}
\begin{center}
\begin{tabular}{| c | c | c | c | c |}
\hline
\textbf{Encoder} & \multicolumn{3}{ c |}{\textbf{BioASQ}}  \\
\cline{2-4}
& \textbf{Epochs} & \textbf{Batch} & \textbf{F1}  \\
\hline

BioBERT v1.1 & 2  & 16 &  0.59  \\ \hline
BioBERT v1.1 & 4 & 16  &  0.65  \\ \hline
BioBERT v1.1 & 8  & 16 &  0.67 \\ \hline
BERT Base & 2 & 16 & 0.59  \\ \hline
BERT Base  & 4 & 16  & 0.55  \\ \hline
BERT Base  & 8 & 16 &  0.69 \\ \hline
Facebook BERT & 2 & 16  & 0.77   \\ \hline
Facebook BERT & 4 & 16  & \textbf{0.81}   \\ \hline
Facebook BERT & 8 & 16 & 0.79 \\ \hline

\end{tabular}
\end{center}
\end{table}

We can observe that Facebook BERT performs better than the other two models since this model has been already fine-tuned on some other question-answering datasets. BioBERT is slightly less performing well than BERT but more stable across epochs. We decided to use the Facebook BERT for our passage retrieval encoders based on the results we've got.

\begin{figure}[ht]
      \includegraphics[width=8cm, height=5cm]{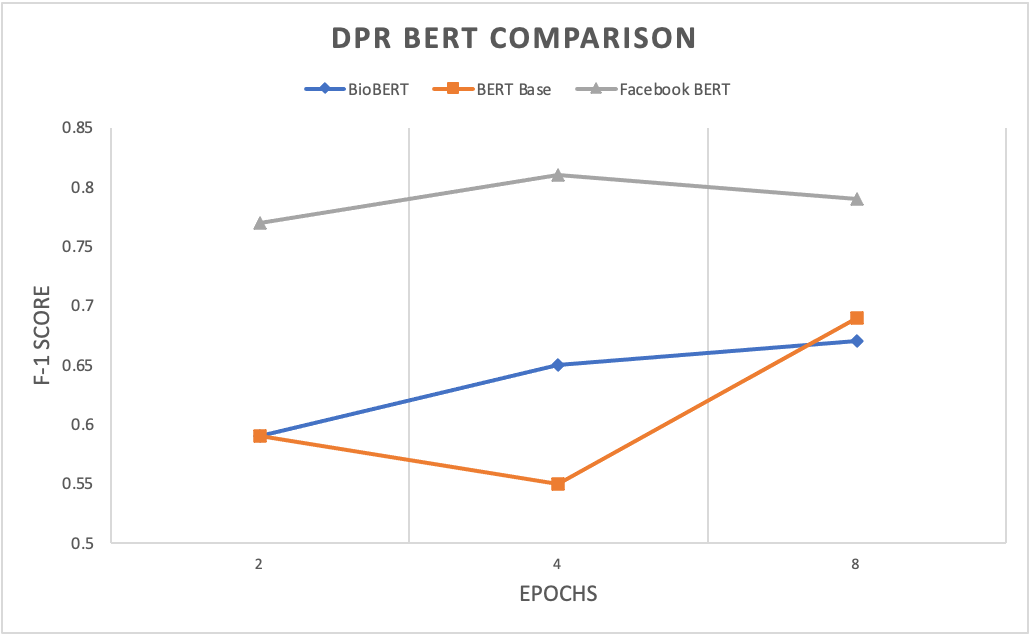}
      \caption{In-Batch Negatives}
\label{Fig:graph}
\end{figure}

It is evident from the graph, Fig. \ref{Fig:graph}, that Facebook BERT has the highest accuracy with respect to BioBERT and BERT-Base. The training is done on 16 batch sizes, for all epochs.
So, that is constant. At epoch 2, it can be observed that BERT-Base and BioBERT perform the same, as is evident from the F-1 score being around 0.6. This is because the BERT base is trained on general data and BioBERT is specifically trained on Biomed data. So, at epoch 2, There is definitely a certain sentiment overlap between the boomed data set and the normal articles. This infers that normal articles have biomed data also. So, it can perform equivalently well in our data for submitting datasets.

As the epoch increases, the variance in the type of data for normal datasets and Biomed-specific datasets increases. So, BERT-BASE starts to fall in performance with a low F-1 score and BioBERT performs exponentially well. However, the anomaly is that, at epoch 3, BioBERT has a less steep growth in performance with respect to BERT-Base. It is worth noting that BERT-Base, given many epochs, can learn and infer patterns better than BioBERT.

\begin{figure}[ht]

      \includegraphics[width=9cm, height=4.5cm]{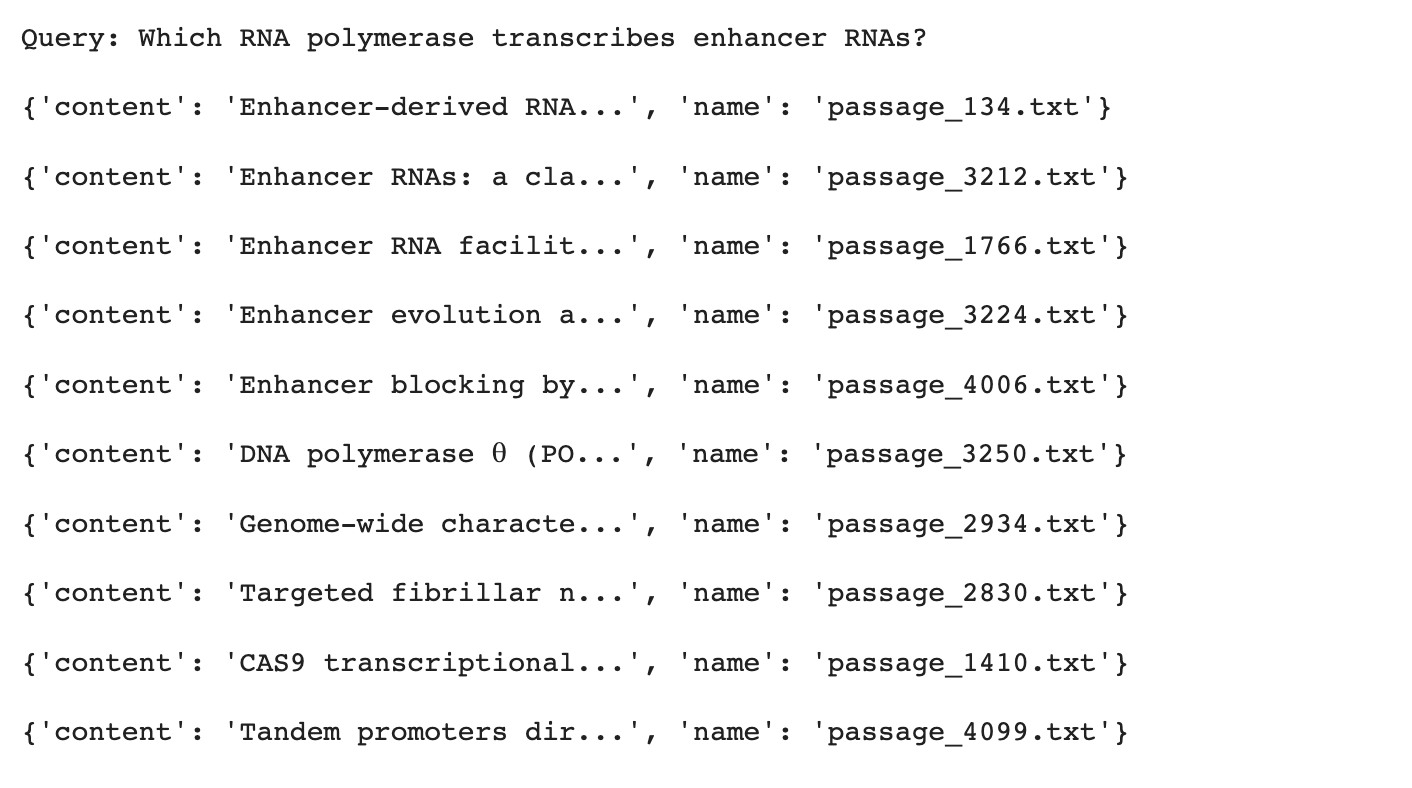}
      \caption{Top 10 documents in response to the question: \textit{"Which RNA polymerase transcribes enhances RNAs?"}}
      \label{Fig:output1}
\end{figure}

\begin{figure}[ht]

      \includegraphics[width=9cm, height=2cm]{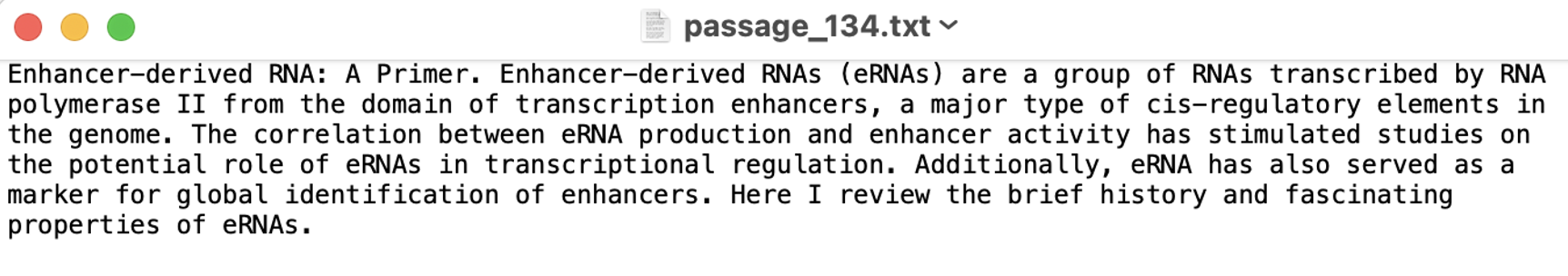}
      \caption{Content of \textit{passage\_134.txt}}
      \label{Fig:output2}
\end{figure}

From Fig. \ref{Fig:output1}, it can be seen that we have tested the query, \textit{"Which RNA polymerase transcribes enhances RNAs?"}. The output is top-10 documents, in which there are possibilities to find the answer. We opened the document \textit{passage\_134.txt}. The content is shown in Fig. \ref{Fig:output2}, which has the correct answer.

\section{Conclusion}

In the present study, we constructed a retriever designed to proficiently extract pertinent PubMed abstracts in response to specific biomedical domain queries. Our approach involved a comparative analysis of various pre-trained language models, each capable of encoding both questions and contexts into high-dimensional vectors. Utilizing Maximum Inner Product Search (MIPS), we targeted the retrieval of the top 10 passages and subsequently assessed the accuracy of these selections. Based on our test dataset, we discerned that the accuracy for the top 10 retrieved passages stands at \textbf{81\%}.

\section{Future Work}

In forthcoming research endeavors, our intention is to augment the volume of training data for the refinement of our model. We are in the process of indexing the entirety of the 30 million articles available on PubMed using FAISS. This will enable us to incorporate hard negatives from the entire corpus of articles. Given the constraints of our current dataset size, we have been limited to retrieving only the top 10 passages. However, with the expanded dataset, we aim to retrieve the top 50 or even 100 pertinent passages. It is imperative that we conduct experiments with varying batch sizes to ascertain their influence on our retrieval scores. Additionally, to further enhance the precision of our results, we are strategizing to develop a reader model that will facilitate the extraction of answers from the given context.


\printbibliography
\end{document}